\documentclass[pdflatex,sn-basic]{sn-jnl}

\usepackage{graphicx}%
\usepackage{amsmath,amssymb,amsfonts}%
\usepackage{amsthm}%
\usepackage{booktabs}%
\usepackage{xspace}%
\usepackage[title]{appendix}%

\newcommand{\xgbcst}{\textsc{XGB-C2ST}\xspace}
\newcommand{\ff}{\textsc{FF}\xspace}          
\newcommand{\tabbyflow}{\textsc{TabbyFlow}\xspace}

\begin{document}

\title[Measuring the Dependency Gap]{Measuring the Dependency Gap:
Diagnosing Inter-Column Fidelity in Tabular Generative Models}

\author*[1]{\fnm{Jie} \sur{Zhang}}\email{stzhangjie@gmail.com}

\affil*[1]{\orgname{Accenture}, \orgaddress{\city{Tokyo}, \postcode{107-8672},
\country{Japan}}}

\abstract{Synthetic tabular data are valued for preserving inter-column dependency, yet
each routine fidelity score is a single number that says neither where that
dependency is lost nor why. We localize the deficit inside a single score. Equipping a classifier two-sample test (C2ST) with a
gradient-boosted discriminator, we decompose it by controlled permutation into
marginal, dependency and numerical--categorical cross components, each read
against a fully-factorized reference that destroys all dependency while leaving
every marginal intact, and against a real-data oracle. The linear detection
score in common use rates that same reference as nearly real, a known weakness
we replicate on four benchmarks. Applied to a flow-matching (TabbyFlow) and a
diffusion (TabDiff) generator, the decomposition exposes a dependency deficit of
the same order in both, and we then narrow its source by elimination: it is not
a structural limit of the mean-field objective, it is not sampling
discretization, and a $16\times$ capacity increase does not close it, while the
same measurement responds sharply when capacity is instead cut eightfold, so the
plateau is measured rather than a blind spot. Destroying dependency outright
collapses minority-class $F_1$ by $0.38$--$0.61$, which bounds what is at stake,
though the generators' much smaller residual deficits do not predict their
remaining shortfalls. What survives points at the objective: nothing in it
scores the joint. Cheap remedies are no substitute: an explicit
cross-coupling module and a post-hoc copula both leave the deficit in place.}

\keywords{Synthetic tabular data, Generative model evaluation,
Classifier two-sample test, Inter-column dependency, Flow matching,
Diffusion models}

\maketitle

\section{Introduction}
\label{sec:intro}

Generative models of tabular data are widely used to share privacy-preserving data \citep{jordon2019pategan,jordon2022synthetic} and augment under-represented classes \citep{dsouza2025overlap,ren2026diffusiongan}. 
Across both applications, the utility of synthetic data hinges on capturing not only marginal distributions but also inter-column dependencies, a requirement particularly critical in financial fraud detection \citep{fiore2019using}, churn prediction \citep{adiputra2024ctganenn}, and clinical risk prognoses \citep{hernandez2022synthetic}.
For instance, neither a young account age nor a large transaction is remarkable alone, yet their co-occurrence often indicates financial fraud. 
Preserving inter-column dependency is therefore pivotal to the overall fidelity of synthetic tabular data and its downstream utility, particularly for minority-class detection \citep{he2009learning}.


Recent state-of-the-art tabular generators are based on flow matching \citep{lipman2023flow,tong2024cfm} and diffusion \citep{ho2020ddpm,song2021score} and adapted to mixed-type tabular data \citep{tabbyflow,tabdiff,cascadedfm}.
A practical question follows immediately: \emph{how do we know a generator preserves dependency?}
Consistent with a known limitation of linear detection scores \citep{hudovernik2024relational}, we find that a degenerate baseline that fits feature marginals separately and samples them independently (i.e., destroying all dependency) still attains near-perfect linear-discriminator classifier two-sample test (C2ST) detection scores ($0.978$–$0.998$, where higher means more indistinguishable from real; equivalently, the discriminator's area under the ROC curve (AUC) is ${\approx}0.5$, at chance) and only mildly degraded pairwise Trend scores ($0.69$–$0.79$).
These results suggest that the linear C2ST primarily certifies marginals rather than dependency, rendering it largely blind to inter-column structure, while pairwise Trend scores are only partially sensitive to it.


To expose what standard metrics miss, we formulate a dependency-aware diagnostic that decomposes C2ST scores.
Equipping C2ST with an interaction-sensitive learner (\xgbcst), we decompose its score into \emph{marginal}, \emph{dependency}, and numerical--categorical \emph{cross} components, each read against two fixed references: a fully-factorized reference (\ff, marginals only, zero dependency) and a real-data oracle.
This framework turns abstract dependency loss into an interpretable, comparative diagnostic for tabular generative models, read under fixed references rather than as an absolute score.


Applied to representative generators, the diagnostic finds a dependency gap in both flow-matching and diffusion models, on datasets where the standard linear C2ST score reports none.
The gap admits three competing explanations: the objective cannot represent joint structure, the model lacks the capacity to learn it, or the objective never asks for it.
We eliminate the first analytically and the second empirically. The second elimination carries a positive control: shrinking the model does move a measurement that enlarging it leaves untouched. We then rule out sampling discretization as a substantial contributor on real data.
What survives is the account that nothing in these objectives ever scores the joint, which points at the objective rather than the architecture as the place to intervene.
Consistent with that reading, no cheap remedy we test closes the gap, though we are explicit about what that null can and cannot support.
Our main contributions are fourfold:

\begin{enumerate}\itemsep2pt
  \item \textbf{A dependency-aware fidelity diagnostic:} We split a \emph{single} \xgbcst\ score into marginal, dependency, and numerical--categorical cross components by controlled permutation, read against established zero-dependency and real-data references \citep{yang2024structured,hudovernik2024relational}; where existing evaluations array separate metrics along a structural spectrum, this localizes the deficit \emph{inside one discriminator's score}, and its cross term reaches mixed numerical--categorical coupling that cumulant-based comparisons cannot (Section~\ref{sec:method}).

  \item \textbf{Empirical characterization of the dependency gap:} A persistent gap appears in both a flow-matching (\tabbyflow) and a diffusion (\textsc{TabDiff}) generator, on benchmarks where the commonly reported linear detection score is blind to inter-column structure, a known weakness \citep{hudovernik2024relational,soberinglook} that we replicate on four datasets. Dependency is \emph{necessary} for minority-class $F_1$, though the generators' residual shortfalls are small and do not track the measured gap (Section~\ref{sec:diagnosis}, Section~\ref{sec:moves}).

  \item \textbf{Narrowing the explanation of the gap:} The gap is not a structural limit of mean-field objectives (exact recovery holds asymptotically \citep{eijkelboom2024vfm}), is not explained by sampling discretization (a step-count sweep on real data), and survives a $16\times$ capacity increase on a model whose measurement does respond to capacity (shrinking width eightfold doubles the gap). This motivates supervising dependency directly in the objective as a next intervention to test (Section~\ref{sec:analysis}, Section~\ref{sec:moves}).

  \item \textbf{Honest negative results:} The residual gap lies in higher-order structure, and a discrete-aware cross-coupling module, a fitted Gaussian copula, and brute-force scaling all fail to close it, a caution against assuming that an added module or more scale will help (Section~\ref{sec:negative}).
\end{enumerate}


\section{Background and Related Work}
\label{sec:background}

\bmhead{Tabular generative models}
Modern tabular synthesizers span generative adversarial networks (GANs) \citep{xu2019ctgan,zhao2021ctabgan}, diffusion models \citep{tabsyn,tabdiff,kotelnikov2023tabddpm,kim2023stasy,lee2023codi}, flow matching \citep{tabbyflow,cascadedfm}, and language models \citep{borisov2023great,liu2023goggle}; for a broader survey, see \citet{borisov2022survey}.
We center our study primarily on \tabbyflow, the tabular instantiation of Exponential-Family Variational Flow Matching (EF-VFM) \citep{tabbyflow}, alongside \textsc{TabDiff} \citep{tabdiff} as a representative mixed-type diffusion generator.
EF-VFM uses a \emph{mean-field} variational posterior that factorizes over columns and a moment-matching objective.
Numerical columns are modeled with a multivariate-Gaussian head whose covariance is isotropic, and categorical columns with per-column categorical heads.
This factorized structure is exactly what makes the question of preserved dependency salient, and motivates our focus.


\bmhead{Evaluating synthetic tabular data}
The dominant fidelity metrics are column-wise and pairwise ``Shape''/``Trend'' similarities together with a detection score, as packaged by widely used libraries such as SDMetrics \citep{patki2016sdv}, whose detection metric trains a logistic-regression discriminator and is provided for single tables. This logistic detection score is widely reported as a fidelity measure, and has been documented as the most common detection-based metric in the relational-synthesis benchmarking literature \citep{hudovernik2024relational}. The latter is a C2ST \citep{gretton2012kernel,lopezpaz2017c2st,kim2021classification}, which casts ``are two samples from the same distribution?'' as a discriminator's accuracy; the linear C2ST we report throughout is this logistic-regression detection score (Section~\ref{sec:method}, Appendix~\ref{app:repro}).
Recent work questions whether these metrics can separate good generators from poor ones \citep{soberinglook,platzer2021holdout}. Two results are closest to our starting point, and we claim neither as a finding of ours. \citet{hudovernik2024relational} shuffle each column of a real table independently, preserving marginals while destroying dependency, and find that a logistic-regression detection score still rates it indistinguishable from real, while a gradient-boosted detector separates it almost perfectly; they trace this to logistic regression's inability to represent column interactions. \citet{soberinglook} reach the same conclusion for single-table generation on an overlapping set of benchmarks and generators: they introduce fully-factorized distributions as baselines, show that these score on par with deep generative models under a logistic-regression C2ST and under Trend, and advocate an XGBoost-based C2ST, which separates the fully-factorized baseline from real data. They additionally propose wNMIS, a mutual-information-based dependency score computed over column pairs. We do not run it, so the empirical relation between the two scores is left open; a pairwise score and a discriminator with joint access to all columns are not interchangeable in principle, as the \textsc{xor} control of Appendix~\ref{app:extra} shows, where every pairwise mutual information is zero while the three-way joint is fully determined. Our fully-factorized reference plays the same role as theirs, though ours resamples each column from its empirical marginal rather than fitting a factorized model, and our discriminator is the one they advocate; our metric-blindness panel (Section~\ref{sec:diagnosis}) should therefore be read as replication, not as a new observation. What we add is what happens next: a decomposition of that single discriminator's score, and an analysis of where the remaining gap comes from.
Our two references are likewise established: \citet{yang2024structured} define exactly this column-wise permutation, together with a real-data half-split, as model-free baselines for reading synthesizers against. That the marginal/joint split matters has been shown from other directions as well \citep{riasat2026dependence,cannon2025assessing}.
We take these observations as our starting point rather than our result. What is new is a \emph{measurement}: one discriminator's score, split into marginal, dependency, and numerical--categorical cross terms by controlled permutation, read against those two baselines, and related to downstream minority-class utility. Existing evaluations array separate metrics along a spectrum of distributional substructure and compare a generator to each in turn; ours localizes the deficit within a single score, and its cross term reaches mixed numerical--categorical coupling, which cumulant-based methods cannot. This is what lets us ask \emph{where} a modern generator loses dependency, and how much it costs, rather than only whether some metric can be fooled.


\bmhead{Modeling dependency}
Copulas separate marginals from dependence \citep{nelsen2006copulas,aas2009vine}, and recent work embeds copula or dependency structure inside generative models.
\citet{pawsterior} adapt variational flow matching to structured \emph{domains} in simulation-based inference (bounded, discrete, and hybrid parameter spaces) rather than to mixed-type tabular generation; their endpoint factorization is itself mean-field, so what differs from our setting is the inference target, not the variational family. \citet{flowcopula} place copulas inside flow and diffusion models, and their construction, which progressively forgets inter-column dependence while holding each marginal fixed, is close in spirit to the fully-factorized reference and the permutation decomposition we use. The roles differ, however: theirs is a model construction, a continuous-time forgetting process used to define a copula-structured generator, whereas ours is a measurement intervention, a one-shot permutation held fixed as a reference. They also work with continuous scientific data rather than mixed-type tables.
Neither targets mixed-type, class-imbalanced tabular generation.
Our work is diagnostic rather than a new generator: we measure where dependency is lost and test whether an in-model cross-coupling mechanism helps.


\bmhead{Mean-field and recovery}
A natural hypothesis is that a mean-field (column-factorized) objective \emph{cannot} represent inter-column dependency.
The variational flow matching literature shows this hypothesis is false in the idealized limit. The mean-field assumption is placed on the \emph{variational} posterior $q_\theta(x_1\mid x_t)$, not on the true posterior; because the conditional vector field is linear in the endpoint, the objective depends on $q_\theta$ only through its per-dimension means, so a fully-factorized approximation is without loss of generality \citep{eijkelboom2024vfm}, and the argument carries over to the exponential-family construction \tabbyflow\ instantiates \citep{tabbyflow}.
Consequently the mean-field optimum recovers the true per-dimension posterior means, the marginal velocity field is exact, and the induced deterministic flow transports to the true \emph{joint} distribution, continuous and categorical alike.
This result is the backbone of our analysis in Section~\ref{sec:analysis}: for EF-VFM, any measured dependency gap must be attributed to finite capacity, optimization, or discretization, not to a structural blind spot. It is established for variational flow matching; Section~\ref{sec:analysis} sets out separately why the same sufficiency structure holds for the diffusion generator we study. Flow matching has also been studied empirically as a tabular synthesizer \citep{vfmrecovery}, which we take as background rather than as the source of this argument.


\section{A Dependency-Aware Fidelity Diagnostic}
\label{sec:method}


\bmhead{Setup}
Let real data $x\sim P$ have $d$ columns partitioned into numerical and categorical blocks, $x=(x^{\mathrm{num}},x^{\mathrm{cat}})$. 
We partition real data into disjoint training and test sets; tabular generators are trained exclusively on the training split to induce a synthetic distribution $Q$.
All fidelity and utility evaluations are conducted against the held-out real test set.
For the downstream utility axis, we follow the train-on-synthetic, test-on-real (TSTR) protocol \citep{esteban2017realvalued}, reporting minority-class $F_1$ of a classifier trained on synthetic data and evaluated on the real test set.


\bmhead{Classifier two-sample test}
For samples $S_P,S_Q$ we form a balanced two-sample set (down-sampling to equal size, so chance accuracy is $0.5$), label real as $1$ and synthetic as $0$, and train a gradient-boosted-tree discriminator \citep{chen2016xgboost} with 5-fold stratified cross-validation.
We report the out-of-fold AUC canonicalized as $\mathrm{c2st}=\max(\mathrm{AUC},1-\mathrm{AUC})$, where $\mathrm{c2st} = 0.5$ indicates indistinguishability (ideal) and $1.0$ complete separability. Because the canonicalization takes a maximum, $\mathrm{c2st}$ is upward-biased under the null: comparing real data against a fresh real subsample lands slightly above $0.5$ (we observe $0.503$--$0.508$), which is why the oracle reference, rather than $0.5$ itself, is the operative floor.
We use gradient-boosted trees rather than a linear model because dependency is carried by high-order column interactions that a linear discriminator cannot capture; this choice is precisely what renders the dependency component visible \citep{lopezpaz2017c2st,soberinglook,grinsztajn2022tree}. Gradient-boosted discriminators have been studied directly as an evaluation instrument for tabular synthetic data \citep{zein2022xgboost}.


\bmhead{Decomposition}
We separate marginal from dependency fidelity by destroying dependency while preserving marginals.
Let $\Pi(\cdot)$ independently permute the rows of \emph{each} column (with independent per-column permutations), so $\Pi(S)$ retains all column marginals but eliminates all inter-column dependency. 
Define
\[
\mathrm{full} = \mathrm{c2st}(S_P, S_Q),\qquad
\mathrm{marg} = \mathrm{c2st}(\Pi(S_P), \Pi(S_Q)),\qquad
\mathrm{dep} = \mathrm{full}-\mathrm{marg}.
\]
$\mathrm{marg}$ isolates discriminability due to mismatched marginals; $\mathrm{dep}$ captures additional discriminability arising from faulty \emph{joint} structure.
This decomposition is \emph{operational} rather than algebraic: its terms are contrasts in discriminator performance under controlled permutation interventions, each obtained by retraining the discriminator on a permuted input, and need not be non-negative or invariant to the discriminator class. Permuting features to test whether a classifier exploits inter-feature dependency is an established device \citep{ojala2010permutation}; here it is applied to a two-sample discriminator rather than to a predictive classifier. We therefore read the components only against the fixed \ff/oracle references below, never as fractions of a statistical divergence.
To isolate mixed-type interactions, we further define a \emph{cross} component via block permutation.
Pairing row $i$'s numerical block $x^{\mathrm{num}}_i$ with row $\pi(i)$'s categorical block $x^{\mathrm{cat}}_{\pi(i)}$ preserves within-block dependency (numerical--numerical and categorical--categorical) while destroying numerical--categorical coupling, yielding $\mathrm{dep}_{\mathrm{cross}} = \mathrm{full} - \mathrm{c2st}(\Pi_{\mathrm{block}}(S_P), \Pi_{\mathrm{block}}(S_Q))$.


\bmhead{Reference points}
We anchor the dependency axis using two real-data references, both adopted from established model-free baselines: \citet{yang2024structured} define a column-wise permutation baseline and a real-data half-split for exactly this bracketing purpose, and use them to identify generators that reproduce column-wise statistics without column dependencies.
The \emph{fully-factorized} reference \ff\ resamples each column independently from its empirical marginal: it preserves marginals but possesses \emph{zero} dependency, and serves as the low-fidelity reference (its dependency component spans the entire deficit).
The \emph{oracle} is a fresh subsample of real training data (retaining correct marginals \emph{and} joint dependency), serving as the high-fidelity reference ($\mathrm{c2st} \approx 0.5$).
These anchor the \emph{dependency-fidelity} axis, not the numerical \xgbcst\ score: on the latter, \ff\ scores highest and the oracle lowest, and neither is a mathematical bound (a generator's empirical score can in principle fall outside the pair). A generator's dependency fidelity is read relative to these two references, computed under the same discriminator.


\bmhead{Protocol and discipline}
Our diagnostic is deterministic given a set of generated samples: references, permutations, and discriminator states are all seeded. 
The real evaluation target is always the held-out test set, never a set used to fit references, preventing circularity. 
We report $\mathrm{mean} \pm \mathrm{std}$ across five random seeds and never select single runs; seeds vary network initialization and data ordering, and, for every generator except \textsc{TabDiff}, sampling noise as well (Appendix~\ref{app:repro}).
Confidence intervals are our primary evidence throughout, with parametric ($t$) $p$-values and sign counts as effect-size summaries; at five seeds significance alone is a blunt instrument, and an interval additionally bounds the effect size, which is what our negative results require (Appendix~\ref{app:repro}).
\xgbcst, the decomposed dependency components, minority-class $F_1$, and the \ff/oracle references are our sole evaluation criteria. Two further metrics appear \emph{only} to demonstrate their diagnostic blindness: the pairwise Trend similarity and a linear-discriminator (logistic-regression) C2ST, the latter as the SDMetrics logistic-detection score, on which $1.0$ means the discriminator is at chance (Appendix~\ref{app:repro}).
Code and configuration files supporting every reported quantity are covered by the Data and Code Availability statement.


\section{What Standard Metrics Miss}
\label{sec:diagnosis}

We apply the diagnostic to \tabbyflow/EF-VFM \citep{tabbyflow} on five class-imbalanced datasets \citep{dua2017uci} (statistics in Appendix~\ref{app:data}): \textsc{adult} and \textsc{default} (the two largest dependency components, our strongest evidence), \textsc{bank} \citep{moro2014bank} (large, categorical-rich, imbalanced), \textsc{magic} (a numerical-only control), and \textsc{shoppers} (qualitative only; smaller dependency effect, below).
For each dataset, we report the generator against the fully-factorized reference (\ff) and the oracle reference over five seeds (Table~\ref{tab:diagnosis}).

\begin{figure}[t]
\centering
\includegraphics[width=0.78\textwidth]{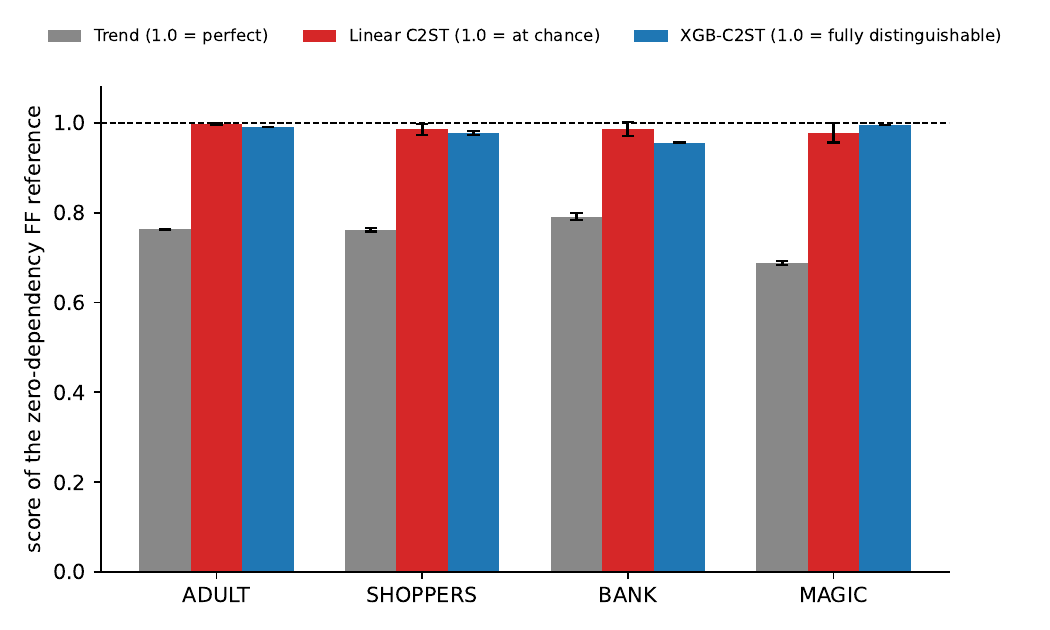}
\caption{Metric blindness. The zero-dependency \ff\ is judged indistinguishable from real by the linear
(logistic-regression) C2ST (detection score ${\approx}1.0$; equivalently the linear discriminator's AUC is
${\approx}0.5$, at chance; see Section~\ref{sec:method}), only mildly penalized by Trend, but correctly judged far
from real by \xgbcst. Trend and the linear C2ST are plotted as quality scores (high = indistinguishable from
real), whereas \xgbcst\ is a discriminator AUC (high = separable = worse). Following their standard usage,
Trend and the linear C2ST are computed against the real \emph{training} set, whereas \xgbcst\ is computed
against the held-out real test set, matching its role as our primary fidelity criterion elsewhere. Bars are
means over five seeds with $\pm1$ std error bars. The panel covers the four datasets
on which the blindness metrics were run; \textsc{default} was not part of this panel}\label{fig:blindness}
\end{figure}

\bmhead{Standard metrics are blind to dependency}
The zero-dependency \ff\ reference, which destroys \emph{all} inter-column structure by construction, is nonetheless judged near-perfect by the linear-discriminator (logistic-regression) C2ST, replicating a known limitation of such scores \citep{hudovernik2024relational,soberinglook}, on each of the four datasets in our metric-blindness panel (\textsc{default} was not part of this panel): detection scores of $0.998_{\pm0.002}$/$0.985_{\pm0.012}$/$0.986_{\pm0.016}$/$0.978_{\pm0.022}$ on \textsc{adult}/\textsc{shoppers}/\textsc{bank}/\textsc{magic}. Equivalently, the linear discriminator sits at chance (AUC ${\approx}0.50$) on data whose inter-column dependency has been entirely destroyed, matching or exceeding the generator's own near-chance score.
Trend is only partially sensitive (across datasets it scores \ff\ at $0.69$--$0.79$ rather than $1.0$), so we rest the blindness claim on the linear C2ST.
A practitioner relying on these metrics would conclude that destroying all inter-column dependency costs almost nothing.
\xgbcst, by contrast, correctly separates \ff\ from real at $0.956$--$0.995$ (Figure~\ref{fig:blindness}): an interaction-sensitive discriminator is needed to see what the linear one misses.
The linear C2ST thus certifies marginals rather than joint structure, and Trend reflects it only partially.

\begin{figure}[t]
\centering
\includegraphics[width=0.78\textwidth]{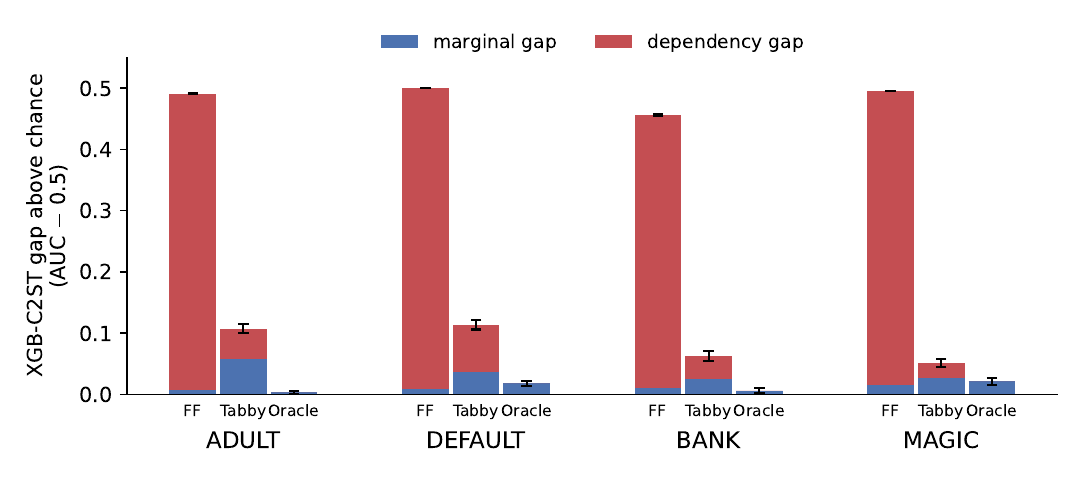}
\caption{The \xgbcst\ gap above chance, decomposed into marginal and dependency segments. Error bars:
$\pm1$ std over five seeds; ``Tabby'' abbreviates \tabbyflow. \ff\ (zero
dependency) is almost all dependency, the oracle is ${\approx}0$, and \tabbyflow\ carries a substantial
dependency segment}\label{fig:gapdecomp}
\end{figure}

\bmhead{A real dependency gap, consistent across datasets}
On \xgbcst, \tabbyflow\ is far from the oracle (e.g., $0.607$ vs.\ $0.503$ on \textsc{adult}), and decomposing the gap attributes a substantial share to inter-column structure (Table~\ref{tab:diagnosis}, Figure~\ref{fig:gapdecomp}). 
The dependency component is significantly positive on all four quantitative datasets, with $95\%$ confidence intervals (CIs) excluding zero (Table~\ref{tab:diagnosis}); the one-sample $t$-tests against zero (five seeds) agree: \textsc{adult} $\mathrm{dep}={+}0.050\pm0.007$ ($p=8.6\times10^{-5}$), \textsc{default} ${+}0.077\pm0.012$ ($p=1.3\times10^{-4}$), \textsc{bank} ${+}0.038\pm0.014$ ($p=0.004$), and the numerical-only control \textsc{magic} ${+}0.025\pm0.016$ ($p=0.023$).
\textsc{shoppers} is directionally consistent (${+}0.017\pm0.016$) but, at the \emph{same} seed-level noise as \textsc{magic} (both $\pm0.016$ over five seeds), its \emph{smaller} dependency component has a $95\%$ confidence interval that includes zero (${+}0.017$, $[{-}0.004,\,{+}0.037]$, five seeds); by the primary-evidence criterion of Section~\ref{sec:method} (confidence intervals, not $p$-values) it is therefore not resolved, so we use it qualitatively only, attributing this to the smaller effect size rather than to its test-set size.
On \textsc{adult}, the component is about half of the total gap above chance ($0.607 - 0.5 = 0.107$). 
The datasets probe different regimes: on \textsc{adult} and \textsc{bank} the gap is dominated by the numerical--categorical cross component ($\mathrm{dep}_{\mathrm{cross}}={+}0.034$ and ${+}0.031$), on \textsc{default} it is carried by within-block structure (numerical--numerical, categorical--categorical, or both; our decomposition does not separate them), and \textsc{magic} shows the gap persists with no categorical columns at all. 
(On \textsc{magic}, which has no categorical columns, the cross component $\mathrm{dep}_{\mathrm{cross}}$ should be zero by construction; the observed values of $-0.004$ and $+0.001$ sit at the noise floor, serving as a sanity check that the block permutation isolates what it claims to.)
Across all datasets, \ff's dependency component absorbs essentially the entire gap ($\mathrm{dep}\approx0.45$--$0.49$), as expected for a reference whose dependency is destroyed by construction.


\bmhead{Dependency is necessary for utility, but the residual shortfall does not track the residual gap}
Training a classifier on synthetic data and evaluating on real test data, the zero-dependency \ff\ collapses to minority-class $F_1$ of $0.102$/$0.087$/$0.016$/$0.242$ on \textsc{adult}/\textsc{default}/\textsc{bank}/\textsc{magic} (standard deviations in Table~\ref{tab:diagnosis}).
The oracle reaches $0.709$/$0.467$/$0.533$/$0.836$, and \tabbyflow\ sits at or just below the oracle ($0.663$/$0.452$/$0.514$/$0.825$).
Two statements should be kept distinct, and only the first is strongly supported.
First, dependency is \emph{necessary} for minority-class utility: destroying it outright, while leaving all marginals intact, makes synthetic data nearly useless for the minority class (the \ff\ collapse, a drop of $0.38$--$0.61$ absolute $F_1$ from the oracle across the four datasets). This bounds the stake of the axis that standard metrics hide.
Second, at the generator's operating point the residual shortfall is small ($0.011$--$0.046$ minority-$F_1$) and, notably, is \emph{not} ordered by the measured dependency component: \textsc{default} carries the largest $\mathrm{dep}$ (${+}0.077$) yet the second-smallest shortfall ($0.015$), and the two within-dataset generator comparisons available (Table~\ref{tab:tabdiff}) point the same way: on \textsc{adult} \textsc{TabDiff} carries the larger $\mathrm{dep}$ and the smaller shortfall, on \textsc{default} \tabbyflow\ does, so in both cases the generator with the larger measured gap has the smaller shortfall, the opposite of what a monotone dependency--utility link would predict.
Moreover, only on \textsc{adult} is the generator-to-oracle shortfall statistically resolvable at five seeds (paired $t$: $\Delta={+}0.046$, $95\%$ CI $[{+}0.039,\,{+}0.052]$); on \textsc{default} ($\Delta={+}0.015$, CI $[{-}0.008,\,{+}0.037]$), \textsc{bank} ($\Delta={+}0.019$, CI $[{-}0.015,\,{+}0.053]$), and \textsc{magic} ($\Delta={+}0.011$, CI $[{-}0.006,\,{+}0.027]$) it is within noise. On all four datasets the paired $\Delta$ equals the difference of the tabulated means, since generator and references use identical five seeds on both sides.
We therefore report the two quantities side by side and do \emph{not} claim that the residual shortfall is attributable to the residual dependency gap.


\begin{table}[t]
\centering
\caption{Dependency-aware diagnosis of \tabbyflow\ against the fully-factorized reference (\ff, zero
dependency) and the real-data oracle reference (5 seeds, mean).
Minority-$F_1$ is shown as mean$_{\pm\text{std}}$. The per-dataset $95\%$ confidence interval on $\mathrm{dep}$
($t$-distribution, four degrees of freedom) excludes zero on all four datasets; these intervals are the
primary evidence, and the one-sample $p$-values are parametric ($t$) effect-size summaries
(see Section~\ref{sec:method}). \xgbcst: $0.5$ = indistinguishable (good), $1.0$ = separable (bad);
lower is better, so \ff\ (highest) and the oracle (lowest) anchor the \emph{fidelity} axis, not the numeric
score. $\mathrm{dep}=\mathrm{full}-\mathrm{marg}$ is the dependency component; $\mathrm{dep}_{\mathrm{cross}}$
isolates numerical--categorical coupling. The generator-to-oracle minority-$F_1$ shortfall is significant only
on \textsc{adult} (paired $t$; Section~\ref{sec:diagnosis}). On \textsc{default} the \ff\ row saturates
($\xgbcst=1.000$), so its $\mathrm{dep}_{\mathrm{cross}}$ of ${+}0.000$ records the ceiling rather than an
absence of numerical--categorical coupling; the column-type attribution in the text is read from the generator
rows only. Metric-blindness values (Trend, linear C2ST) are in the
text and Figure~\ref{fig:blindness}.}
\label{tab:diagnosis}
\begin{tabular}{lcccc}
\toprule
Variant & \xgbcst & dep & $\mathrm{dep}_{\mathrm{cross}}$ & min.\ $F_1$ \\
\midrule
\multicolumn{5}{l}{\emph{\textsc{adult}} ($\mathrm{dep}>0$; $95\%$ CI excludes zero, $p=8.6\times10^{-5}$)}\\
\quad \ff\ (zero-dep.\ ref.) & $0.991$ & ${+}0.484$ & ${+}0.054$ & $0.102_{\pm 0.004}$ \\
\quad \tabbyflow         & $0.607$ & ${+}0.050$ & ${+}0.034$ & $0.663_{\pm 0.006}$ \\
\quad oracle (real-data ref.) & $0.503$ & $0.000$ & ${+}0.001$ & $0.709_{\pm 0.005}$ \\
\midrule
\multicolumn{5}{l}{\emph{\textsc{default}} ($\mathrm{dep}>0$; $95\%$ CI excludes zero, $p=1.3\times10^{-4}$)}\\
\quad \ff\ (zero-dep.\ ref.) & $1.000$ & ${+}0.491$ & ${+}0.000$ & $0.087_{\pm 0.040}$ \\
\quad \tabbyflow         & $0.613$ & ${+}0.077$ & ${+}0.010$ & $0.452_{\pm 0.007}$ \\
\quad oracle (real-data ref.) & $0.506$ & ${-}0.013$ & ${+}0.003$ & $0.467_{\pm 0.018}$ \\
\midrule
\multicolumn{5}{l}{\emph{\textsc{bank}} ($\mathrm{dep}>0$; $95\%$ CI excludes zero, $p=0.004$)}\\
\quad \ff\ (zero-dep.\ ref.) & $0.956$ & ${+}0.446$ & ${+}0.073$ & $0.016_{\pm 0.005}$ \\
\quad \tabbyflow         & $0.562$ & ${+}0.038$ & ${+}0.031$ & $0.514_{\pm 0.018}$ \\
\quad oracle (real-data ref.) & $0.504$ & ${-}0.002$ & ${-}0.004$ & $0.533_{\pm 0.017}$ \\
\midrule
\multicolumn{5}{l}{\emph{\textsc{magic}} (numerical-only control; $\mathrm{dep}>0$; $95\%$ CI excludes zero, $p=0.023$)}\\
\quad \ff\ (zero-dep.\ ref.) & $0.995$ & ${+}0.479$ & ${+}0.001$ & $0.242_{\pm 0.040}$ \\
\quad \tabbyflow         & $0.551$ & ${+}0.025$ & ${-}0.004$ & $0.825_{\pm 0.006}$ \\
\quad oracle (real-data ref.) & $0.508$ & ${-}0.013$ & ${+}0.001$ & $0.836_{\pm 0.008}$ \\
\botrule
\end{tabular}
\end{table}


\section{Is the Gap a Structural Blind Spot? No}
\label{sec:analysis}

The dependency gap in Section~\ref{sec:diagnosis} admits three competing explanations. 
\textbf{(H1)}~The mean-field objective is structurally unable to recover inter-column dependency: its optimum loses it. 
\textbf{(H2)}~It is recoverable, but the baseline lacks the \emph{capacity} to do so. 
\textbf{(H3)}~The dependency is recoverable and the baseline has ample capacity, but at finite capacity and a practical optimization budget the \emph{objective} exerts no gradient pressure toward joint fidelity, so the available capacity is not allocated to it.
We rule out H1 in this section, with a recovery result for variational flow matching. Section~\ref{sec:moves} then finds no effect from scaling the model $16\times$ under matched optimization (H2, on the datasets where the test has power) and no substantial contribution from sampling discretization, leaving the account that the deficit lies in what the objective rewards rather than in what the architecture can represent or how large it is. We are explicit there about what that leaves: H3 rests on elimination, and no experiment here adds dependency supervision and watches the gap close.
The distinction is practical because H2 and H3 differ in what scaling should do, and Section~\ref{sec:moves} separates them: the gap survives a $16\times$ increase, while the same measurement responds sharply when capacity is cut instead.


\bmhead{The objective supervises only per-dimension posterior means}
EF-VFM trains a factorized variational posterior $q_\theta(x_1\mid x_t)=\prod_i q_{\theta,i}(x_1^i\mid x_t)$ against a moment-matching objective that decomposes over columns. 
Two consequences follow. 
First, the objective directly supervises each column's \emph{per-dimension} posterior mean $\mathbb{E}[x_1^i\mid x_t]$ and nothing about the joint over $x_1$. 
Second, the joint dependency of the \emph{generated} samples is therefore not directly supervised: it is an emergent property of integrating the induced velocity field, recovered only insofar as those per-dimension means are learned accurately. 
We state this as an observation rather than a hard theorem; it is the lens through which the rest of the section reads.


\bmhead{No structural blind spot}
Crucially, accurate per-dimension means are \emph{sufficient}. 
Because the interpolation path is affine in the (one-hot-encoded) endpoint, the marginal velocity field $v^\star(x_t,t)=\mathbb{E}[u_t'(x_1)\mid x_t]$ depends only on $\mathbb{E}[x_1\mid x_t]$.
The mean-field optimum attains these per-dimension means, so $v_\theta=v^\star$, and the deterministic flow following $v^\star$ transports to the true \emph{joint} distribution, for continuous and categorical columns alike, in the infinite-capacity limit \citep{eijkelboom2024vfm,tabbyflow}. 
The dependency, though never explicitly supervised, is reconstructed dynamically through the shared dependence of each column's velocity on the full state $x_t$. 
There is therefore no structural blind spot for EF-VFM, and any observed gap must be attributed to finite capacity, optimization, or sampling discretization; the objective's \emph{asymptotic capability} is not at issue.
The same sufficiency structure carries over to the diffusion generator we study, by standard results rather than any new argument of ours; Appendix~\ref{app:diffusion} sets it out. The one contribution it does not cover, for either generator, is sampling discretization at a finite step count, which we therefore probe directly below.
One clarification reconciles this with H3 below: exactness \emph{at} the optimum does not imply gradient pressure \emph{toward} dependence at finite capacity. H3 is the claim that this pressure is absent, so it lives \emph{within} the finite-capacity optimization category rather than beside it: a specific mechanism (no gradient signal for the joint), not a fourth alternative to capacity and optimization.


One worry is specific to categorical generation: EF-VFM decodes each categorical column independently at the final step, which could discard joint categorical structure. A controlled task addresses it, in which one categorical variable is the exclusive-or (XOR) of two others, so that all pairwise correlations vanish and only the three-way joint carries information. The model recovers that joint despite independent per-column decoding (Appendix~\ref{app:extra}), consistent with the argument above.


\section{What Moves the Gap, and What Does Not}
\label{sec:moves}

Section~\ref{sec:analysis} establishes that the gap need not be there. This section asks what does and does not move it: model capacity, swept in both directions; the number of sampling steps; and a change of generator across the flow-matching/diffusion boundary.


\bmhead{Ruling out insufficient capacity (H2)}
The recovery guarantee is asymptotic; it does not say the gap closes at any \emph{practical} capacity. 
The insufficient-capacity account (H2) predicts that scaling the model up should shrink the dependency component. 
We test it with a width sweep, everything else fixed ($\mathrm{dim}_t$, the network width, $\in\{1024,2048,4096\}$, i.e., $1{\times}/2{\times}/4{\times}$, giving $10.6$M/$42$M/$168$M parameters, so the $4\times$ width has $\approx\!16\times$ the parameters), five seeds each, on the three datasets whose large models train cleanly (each seed's final loss reaches the small model's level: \textsc{adult} $11.7$--$11.9$, \textsc{bank} $\approx\!13.0$, \textsc{magic} $\approx\!8.9$; no bimodality). This gate screens for optimization \emph{failure}; as the downward sweep below shows, the training loss is nearly flat across a $57\times$ range of capacity and so cannot by itself certify that a larger model is trained to its capacity. On \textsc{default} the plateau scheduler collapses the learning rate at every width, so the matched-optimization premise cannot be verified and that dataset is excluded from the capacity comparison rather than interpreted (Appendix~\ref{app:repro}). 
The dependency component across $1{\times}/2{\times}/4{\times}$ is $0.050/0.051/0.058$ on \textsc{adult}, $0.038/0.041/0.043$ on \textsc{bank}, and $0.025/0.016/0.015$ on \textsc{magic}.
Paired $t$-tests of $4{\times}$ against $1{\times}$ (same seeds) find no significant change on any dataset ($p=0.098/0.484/0.185$), and per-seed linear trends across the three widths agree (Appendix~\ref{app:extra}).
On \textsc{adult}, our highest-signal cell, the direction is flat-to-slightly-up and the $95\%$ CI on the $4{\times}{-}1{\times}$ difference, $[-0.002,\,+0.019]$, excludes any true reduction larger than $0.0024$ (about $5\%$ of its $0.050$ component); \textsc{bank} is consistent but wider ($[-0.013,\,+0.022]$). \textsc{magic}, by contrast, points the other way, with an interval too wide to constrain anything (Appendix~\ref{app:extra}); we therefore rest the capacity conclusion on \textsc{adult} and, more weakly, \textsc{bank}, and do not claim it holds on \textsc{magic}.
We claim only that a $16\times$ increase does not close the gap where the sweep trains cleanly and the test has power, not that the gap is capacity-independent in general; a finite sweep also cannot contradict the asymptotic recovery limit, so none of this resurrects H1.


\bmhead{The measurement does respond to capacity (downward sweep)}
The sweep above scales \emph{up} from a $1{\times}$ model that already carries $10.6$M parameters for ${\sim}49$k \textsc{adult} rows. A null from such a sweep is ambiguous: it is equally consistent with the gap being insensitive to capacity and with $\mathrm{dep}$ being unable to detect capacity effects at all. We therefore run the sweep in the other direction on \textsc{adult}, at $\mathrm{dim}_t\in\{128,256,512\}$ ($\tfrac{1}{8}{\times}/\tfrac{1}{4}{\times}/\tfrac{1}{2}{\times}$), five seeds each, everything else fixed.
The measurement responds sharply. The dependency component rises monotonically as width falls ($0.050/0.060/0.074/0.099$ at $1{\times}/\tfrac12{\times}/\tfrac14{\times}/\tfrac18{\times}$), and every paired contrast against $1{\times}$ is significant ($p=0.0023/0.0003/0.0005$). The $1{\times}$ model therefore sits at the knee of a curve the diagnostic can trace, and the flat region above it is a measured plateau rather than a blind spot: this is the positive control the upward sweep alone cannot supply.
The two components saturate in a revealing order. The marginal component is already at its floor by $\tfrac12{\times}$ ($\Delta$ against $1{\times}$: $-0.0003$, CI $[-0.006,\,+0.006]$, $p=0.91$), while over the \emph{same} width doubling the dependency component still improves significantly ($\Delta={+}0.010$, CI $[+0.006,\,+0.014]$, $p=0.0023$). Dependency is the last thing capacity buys. Judging sufficiency by marginal quality, or by the training loss (which varies only over $11.84$--$12.10$ across this entire $57\times$ range of parameter count), would stop at a width where dependency is still being learned.
Downstream utility, by contrast, does not follow the dependency component at all. Doubling $\mathrm{dep}$ (from $0.050$ at $1{\times}$ to $0.099$ at $\tfrac18{\times}$) leaves minority-class $F_1$ unchanged ($\Delta={+}0.002$, $95\%$ CI $[-0.005,\,+0.010]$), excluding any true change larger than about $0.010$ absolute $F_1$: roughly $1.6\%$ of the \ff\ collapse on this dataset and a fifth of the generator's own residual shortfall. This is a within-dataset, seed-matched version of the dissociation reported in Section~\ref{sec:diagnosis}, and it sharpens the reading there: utility responds to dependency with a threshold, collapsing when dependency is destroyed outright but flat over the range in which these generators actually operate. One of the twenty runs is an $F_1$ outlier, so this quantity has heavier seed-level tails than the within-width standard deviations suggest; the contrast above contains no such point (Appendix~\ref{app:extra}).


\bmhead{Sampling discretization is a minor contributor (step-count sweep)}
Discretization is entangled with H3 in the analysis above, and the recovery argument covers both generators only in the continuous-time limit. We therefore sweep the sampling step count (number of function evaluations, NFE) directly on the real datasets, resampling from the trained checkpoints so that nothing else changes, five seeds per setting.
For EF-VFM the gap is essentially step-independent on both \textsc{adult} and \textsc{default}: pushing the step count far above the default changes $\mathrm{dep}$ by ${-}2.7\%$ and ${+}2.0\%$ of its value, neither resolvable. \textsc{TabDiff}, whose categorical columns use masked diffusion and therefore decode several positions in parallel at finite step counts, does show a resolvable effect on \textsc{adult} ($p=0.0003$), but a bounded one: raising the step count tenfold reduces $\mathrm{dep}$ from $0.051$ to $0.045$, about a tenth of the gap, and leaves the rest (Appendix~\ref{app:repro}). On \textsc{default} the effect is not resolvable.
Discretization is thus a real but minor contributor for the generator where the mechanism predicts it, and negligible for the other.
What remains is consistent with H3: the gap is unmoved by more capacity on the datasets where that test has power, as expected if the objective provides no gradient signal to allocate capacity to dependency. We are explicit about the epistemic status. H3 is supported by elimination, not by a direct intervention that adds dependency supervision and closes the gap; and it is one specific mechanism inside the finite-capacity optimization category, which we do not separate from other under-optimization of the kind documented for \textsc{default} in Appendix~\ref{app:repro}. H1--H3 are therefore not exhaustive.


\bmhead{The gap replicates on a second generator across the flow/diffusion boundary}
The analysis so far concerns one objective.
There is a concrete, testable reason to expect the gap elsewhere: \textsc{TabDiff}'s denoising objective, like EF-VFM's moment-matching objective, decomposes additively over columns and therefore supervises only per-dimension conditionals, placing no term directly on the joint. H3 thus \emph{predicts} a gap wherever this additive-over-columns structure holds, not just in EF-VFM. This is a mechanistic prediction, not an appeal to generators in general.
To test it, we run the \emph{identical} diagnostic (same data splits, same \ff/oracle references, same discriminator) on \textsc{TabDiff} \citep{tabdiff}, a mixed-type \emph{diffusion} model, crossing the diffusion/flow-matching boundary; we first verify that its reported Shape-similarity and machine-learning-efficacy scores reproduce, so the measured gap is not an artifact of under-training (Appendix~\ref{app:repro}). 
On \textsc{adult} and \textsc{default} (five seeds each), \textsc{TabDiff} shows a significant dependency gap of the same order as \tabbyflow\ (Table~\ref{tab:tabdiff}): $\mathrm{dep}={+}0.054$ (95\% CI $[0.049,\,0.059]$, $p=8.5\times10^{-6}$) on \textsc{adult} and ${+}0.055$ (95\% CI $[0.042,\,0.068]$, $p=3.4\times10^{-4}$) on \textsc{default}, with the same ordering (\ff\ $>$ generator $>$ oracle) and a comparable minority-$F_1$ shortfall. 
The similarity is not an artifact of one discriminator: across three discriminator strengths, \textsc{TabDiff}'s gap tracks \tabbyflow's at every setting and the \ff{}$\,>\,$generator$\,>\,$oracle ordering is preserved throughout (Appendix~\ref{app:extra}).
The absolute magnitude does grow with discriminator strength, which is exactly why we read $\mathrm{dep}$ against same-discriminator \ff/oracle references rather than as an absolute quantity. 
Two generators from different paradigms thus leave dependency gaps of the same order, consistent with the mechanistic prediction above: the deficit tracks what these objectives \emph{share} (additive-over-columns supervision), not anything specific to EF-VFM. We read this as replication on a second generator on two datasets, not as a claim about tabular generators in general.


\begin{table}[t]
\centering
\caption{The dependency gap replicates on a second generator across the diffusion/flow-matching boundary.
\tabbyflow\ (flow matching) versus \textsc{TabDiff} (diffusion) under the identical diagnostic (same
splits, \ff/oracle references, and discriminator); five seeds. We report the dependency component
$\mathrm{dep}$ and its numerical--categorical cross subcomponent $\mathrm{dep}_{\mathrm{cross}}$ for both
generators; \textsc{TabDiff}'s $\mathrm{dep}$ carries a one-sample $95\%$ CI against zero ($t$-distribution,
four degrees of freedom; $p=8.5\times10^{-6}$ on \textsc{adult}, $p=3.4\times10^{-4}$ on \textsc{default}). While $\mathrm{dep}$
is of the same order for both generators, the ordering of $\mathrm{dep}_{\mathrm{cross}}$ flips between
datasets (Section~\ref{sec:discussion}). The final column is \textsc{TabDiff}'s minority-class $F_1$, to be
read against \tabbyflow's in Table~\ref{tab:diagnosis}.}
\label{tab:tabdiff}
\begin{tabular}{lcccccc}
\toprule
& \multicolumn{2}{c}{$\mathrm{dep}$} & \multicolumn{2}{c}{$\mathrm{dep}_{\mathrm{cross}}$} & \multicolumn{2}{c}{\textsc{TabDiff}} \\
\cmidrule(lr){2-3}\cmidrule(lr){4-5}\cmidrule(lr){6-7}
Dataset & \tabbyflow & \textsc{TabDiff} & \tabbyflow & \textsc{TabDiff} & 95\% CI & min.\ $F_1$ \\
\midrule
\textsc{adult}   & ${+}0.050$ & ${+}0.054$ & ${+}0.034$ & ${+}0.029$ & $[0.049,\,0.059]$ & $0.671$ \\
\textsc{default} & ${+}0.077$ & ${+}0.055$ & ${+}0.010$ & ${+}0.016$ & $[0.042,\,0.068]$ & $0.440$ \\
\botrule
\end{tabular}
\end{table}


\section{Can In-Model Fixes Close It?}
\label{sec:negative}


If the deficit lies in what the objective supervises, then changing the architecture, or patching the samples afterwards, should not help. That is a prediction about cheap remedies rather than a test of the account itself, and this section reports the ones we tried.


\bmhead{An explicit in-model dependency mechanism}
Guided by the column-type attribution of the gap (on \textsc{adult} the dependency is dominated by the numerical--categorical cross component), we augment the generator with a discrete-aware cross-coupling head: a pooled summary of the decoder's numerical-column representations is mapped through a single linear layer and added to every categorical logit, so that the final categorical head, which is otherwise per-column, carries an explicit cross-column term. 
We verify, via a sampler-level check, that the added mechanism actually enters the generated samples (and is not merely a training-loss term). 
We then compare the modified model against the unmodified baseline using a \emph{paired} protocol: for each of five matched seeds we form per-seed differences $\Delta_i = \text{fix}_i - \text{base}_i$ and report $\mathrm{mean} \pm \mathrm{std}$ and sign counts, which removes shared initialization and sampling noise. 
We prespecify the success criteria: a significant decrease in the targeted (numerical--categorical cross) dependency component, a significant increase in minority $F_1$, and a genuine decrease in full \xgbcst.
The intervention meets none of them on any of the three datasets: on \textsc{adult} the targeted component moves by $\Delta\mathrm{dep}_{\mathrm{cross}}={+}0.002\pm0.003$, in the wrong direction, with a $95\%$ interval of $[-0.001,\,+0.005]$ that excludes any true improvement larger than about $3\%$ of that component; \textsc{bank} and \textsc{default} are consistent but wider. Minority $F_1$ and full \xgbcst\ are likewise unchanged. Per-dataset numbers, intervals, and the sampler-level check are in Appendix~\ref{app:negative}.
Two points about scope carry the reading. The head does not change the mean-field factorization of $q_\theta(x_1\mid x_t)$ over columns, nor does it enlarge the function class, since the baseline's encoder and decoder are joint self-attention modules; what it changes is the inductive bias, supplying the coupling along an explicit additive path rather than requiring the network to learn the routing. And the path is narrow, a four-dimensional pooled summary mapped onto all categorical logits, so it is rank-limited by construction. That is on its own a sufficient explanation of the null, which means the null is \emph{over-determined}: we cannot separate ``the objective is the binding constraint'' from ``this pathway was too narrow to matter.'' We therefore report it as a result about \emph{remedies}, not as evidence about the objective, and we do not count it among the grounds for H3 in Section~\ref{sec:moves}. What it does establish is practical: a cheap, targeted architectural patch aimed at the component the diagnostic localizes does not move it.



\bmhead{The residual gap is higher-order; post-hoc and resampling routes do not supply it}
Three observations clarify why these fixes fail. 
First, the residual gap is \emph{higher-order}: the generator already matches the second-order numerical--categorical cross-moments of the real data to near the oracle floor (a mean standardized moment gap of $0.067$/$0.042$/$0.053$ on \textsc{adult}/\textsc{default}/\textsc{bank}, versus $0.052$/$0.033$/$0.051$ for the oracle and $0.248$/$0.196$/$0.173$ for the zero-dependency \ff; five seeds), leaving under a tenth of the \ff-to-oracle range on each of the three datasets where this cross-moment is defined (\textsc{magic} has no categorical columns, so it does not apply there). 
A second-moment or covariance-matching fix therefore has essentially nothing to correct: the mismatch lives in higher-order structure that such a term does not see. This connects to two prior findings. \citet{cannon2025assessing} report that a GAN and a fine-tuned language model reproduce marginals while failing to reproduce third- and fourth-order joint cumulants, with the notable exception of the GAN on their preprocessing of \textsc{adult}; and \citet{riasat2026dependence} propose covariance-level dependence fidelity as an evaluation criterion, showing that dependence divergence alone can reverse the sign of population regression coefficients. Here that covariance-level criterion is already met, so the deficit sits strictly above it. The downstream consequence we measure is smaller than their population-level construction would suggest, a finite-sample dissociation rather than a contradiction. Moreover, joint cumulants are undefined on categorical columns, so on a schema like \textsc{adult}'s a cumulant-based comparison sees only the numerical columns, whereas our cross component is defined precisely on the numerical--categorical coupling such comparisons cannot reach. We accordingly do not implement one, and state the reason explicitly: the diagnostic already localizes the residual gap above the second order (the standardized second-moment gap sits near the oracle floor), so a covariance term is predicted \emph{a priori} to be inert, and a null result from it would be uninformative. A useful positive intervention would instead have to supply \emph{higher-order} dependency (e.g., a joint or interaction-sensitive term); testing such a term is the natural next step this diagnosis motivates but does not itself carry out. 
The same logic bounds invertible-preprocessing strategies more broadly: linear whitening (e.g., principal component analysis, PCA) removes only second-order correlation, and rule-based pipelines that strip exact functional or hierarchical constraints before generation \citep{long2025llmtablogic} remove only deterministic structure, and both restore what they remove by construction, but neither supplies the higher-order statistical dependency that constitutes the gap.
Second, a fitted Gaussian copula, the standard route for imposing dependency on a set of marginals, does not supply the missing structure on these datasets: our discriminator separates its samples at the $\approx\!1.0$ ceiling of \xgbcst. Third, a resampling baseline (SMOTE \citep{chawla2002smote}) reaches minority-class $F_1$ comparable to training on real data, but by reusing real rows rather than synthesizing the joint, so it does not speak to the generative-dependency question. Appendix~\ref{app:negative} gives both in full, including what the copula result does and does not license.
%


\bmhead{Takeaway}
None of the cheap interventions we test closes the gap.
The discrete-aware cross-coupling module we test does not, though its null is over-determined and we read it as a caution about cheap patches rather than as evidence about the objective (above); the residual gap is higher-order, so a covariance-matching fix has nothing to correct; the post-hoc Gaussian copula we test does not supply the missing structure on the datasets we run it on; and, per Section~\ref{sec:moves}, scaling capacity $16\times$ leaves the gap essentially untouched on \textsc{adult} and (more weakly) \textsc{bank}, with \textsc{magic} uninformative.
This is consistent with dependency being unsupervised by the objective, though it does not by itself establish that.


\section{Discussion and Limitations}
\label{sec:discussion}


\bmhead{What we claim, and how strongly}
Two qualitative findings hold across datasets: metric blindness (a known limitation \citep{hudovernik2024relational,soberinglook}, replicated here), and the absence of a structural blind spot (which, for EF-VFM, follows from the recovery argument of \citet{eijkelboom2024vfm} and the XOR control).
The dependency$\leftrightarrow$utility link holds in only one direction: dependency is \emph{necessary} for minority-class utility (the \ff\ collapse, robust across datasets), but the generators' residual shortfall does \emph{not} track their residual dependency gap: the shortfall itself is statistically resolvable only on \textsc{adult}, and the within-dataset cross-generator ordering inverts (Section~\ref{sec:diagnosis}).
The quantitative dependency gap is significant on all four quantitative datasets ($95\%$ CIs excluding zero; one-sample $t$, $p\le0.023$), strongest on \textsc{adult} and \textsc{default}, which we treat as the primary evidence, with \textsc{bank}/\textsc{magic} as replication and control and \textsc{shoppers} qualitative only (smaller effect size, not separable from zero at the observed noise; Section~\ref{sec:diagnosis}).
Our capacity finding rests on width sweeps on \textsc{adult}, \textsc{bank}, and \textsc{magic} where the larger models converge cleanly (paired $t$, all $p>0.05$), and is carried mainly by \textsc{adult} (\textsc{magic}'s interval is uninformative and points the other way; Section~\ref{sec:moves}). The downward sweep that supplies the positive control for this null, together with the saturation-order and utility results read off it, was run on \textsc{adult} only, and we do not claim either generalizes to the other datasets.
Trend is partially sensitive to destroyed dependency (scoring the \ff\ reference $0.69$--$0.79$ across datasets), so the blindness claim rests on the linear C2ST.
Our datasets are standard public benchmarks; although \textsc{bank} and \textsc{default} are financial, validation on production-scale financial or clinical data is left to future work.
The second-generator check covers two generators on two datasets: both show a significant dependency gap of the same order, but which generator's gap is larger is not consistent across the two datasets, and the numerical--categorical cross subcomponent is likewise not consistent across generators.
We therefore read the cross-generator evidence at the level of the aggregate dependency component, not its subcomponents.


\bmhead{Approximations and what the diagnostic does not establish}
As defined in Section~\ref{sec:method}, the decomposition is \emph{operational}, not algebraic: its components are contrasts in \xgbcst\ performance under controlled permutations, so a ``share'' is a discriminability contrast rather than a fraction of a statistical divergence, need not be non-negative, and depends on discriminator capacity.
Every component is therefore anchored to the \ff\ and oracle references computed under the \emph{same} discriminator, and carries no meaning apart from them. The numerical--categorical cross component is isolated by a block permutation that destroys cross-coupling while preserving within-block structure.
The diagnostic is also not calibrated against a known ground truth: we show that its components separate a zero-dependency reference from real data and preserve the \ff{}${>}$generator${>}$oracle ordering across discriminator strengths (Appendix~\ref{app:extra}), but we do not establish how $\mathrm{dep}$ scales with a controlled quantity of destroyed dependency. Its magnitude therefore has no interpretable units beyond the two references, and a calibration curve is left to future work.


\bmhead{Guidance for practitioners and method designers}
For practitioners evaluating tabular synthesizers: report the dependency decomposition with fully-factorized and oracle references, and do not rely on Trend or the linear (logistic-regression) C2ST to certify dependency, since a zero-dependency baseline passes them. This reinforces an existing recommendation to prefer tree-ensemble discriminators for detection-based fidelity \citep{zein2022xgboost,hudovernik2024relational,soberinglook}. The case for measuring it is not that the size of a generator's residual gap predicts its downstream cost, which across our datasets it does not (Section~\ref{sec:diagnosis}), but that the axis is otherwise uncertified, and the \ff\ collapse bounds what is at stake should a generator degrade along it. 
For method designers: a $16\times$ capacity increase did not close the gap on the datasets where the sweep trains cleanly, carried by \textsc{adult}, and the cheap architectural patch we tried did not either, though the latter null is over-determined (Section~\ref{sec:negative}). This is consistent with what the objective supervises being the binding constraint, though we do not establish it, so a natural next step to test is objectives that reward dependency directly.
Finally, an open question left by our analysis is to characterize \emph{which} dependency structures are hard to recover and at what rate; we do not give such a characterization here.




\backmatter

\bmhead{Acknowledgements}
The author thanks the maintainers of the UCI Machine Learning Repository and
the authors of TabbyFlow/EF-VFM and TabDiff for releasing their code.

\section*{Statements and Declarations}

\bmhead{Funding}
No funding, grants, or other support was received for conducting this study or
preparing this manuscript.

\bmhead{Competing interests}
The author is employed by Accenture. This work is the author's own independent
research, carried out on the author's own initiative and resources and not
funded or sponsored by Accenture; it uses only publicly available benchmark
datasets from the UCI Machine Learning Repository and no proprietary or client
data. The author has no other relevant financial or non-financial interests to
disclose.

\bmhead{Ethics approval and consent to participate}
Not applicable. This work does not involve experiments with human participants
or animals. All datasets are publicly available, de-identified benchmark
tables distributed by the UCI Machine Learning Repository.

\bmhead{Consent for publication}
Not applicable.

\bmhead{Data availability}
No new empirical data were collected. All five datasets analyzed in this study
are publicly available from the UCI Machine Learning Repository
(\url{http://archive.ics.uci.edu/ml}): \textsc{adult}, \textsc{default} (Default
of Credit Card Clients), \textsc{bank} (Bank Marketing), \textsc{magic} (MAGIC
Gamma Telescope) and \textsc{shoppers} (Online Shoppers Purchasing Intention).
Column-type partitions and held-out test-set sizes are documented in
Appendix~\ref{app:data}. The synthetic
tables analyzed here are produced by the code covered below and can be
regenerated from the recorded seeds.

\bmhead{Code availability}
The complete codebase, configuration files and the exact seeds used for every
reported quantity will be deposited in a public repository and released upon
publication; a persistent identifier will be provided at that time. The
generators evaluated here (TabbyFlow/EF-VFM and TabDiff) are used with their
authors' released code and default recipes, as described in
Appendix~\ref{app:repro}.

\bmhead{Author contributions}
J.Z.\ is the sole author and carried out all aspects of the work reported
here.


\bibliography{references}


\begin{appendices}

\section{Dataset statistics}
\label{app:data}

Table~\ref{tab:datasets} summarizes the datasets. All are standard public, class-imbalanced binary-classification
benchmarks. Column-type partitions and the held-out test sizes are those used by our pipeline; \textsc{adult}
uses its canonical train/test split, and the others use a fixed held-out test set (re-used across all seeds so
that the real evaluation target is never refit). \textsc{magic} has no categorical columns and serves as a
numerical-only control. \textsc{shoppers} is used qualitatively only because its dependency component, at the same seed-level noise as \textsc{magic}, is smaller and does not separate from zero (Section~\ref{sec:diagnosis}).

\begin{table}[h]
\centering
\caption{Dataset statistics. ``min.\ frac.'' is the minority-class fraction over the full dataset;
$n_{\text{test}}$ is the held-out test-set size (the real evaluation target). Counts are computed directly
from the data.}
\label{tab:datasets}
\begin{tabular}{@{}lccccccc@{}}
\toprule
Dataset & rows & \#num & \#cat & \#features & min.\ frac. & $n_{\text{test}}$ & role \\
\midrule
\textsc{adult}    & $48{,}842$ & $6$  & $8$ & $14$ & $0.239$ & $16{,}281$ & primary \\
\textsc{default}  & $30{,}000$ & $14$ & $9$ & $23$ & $0.221$ & $3{,}000$  & primary \\
\textsc{bank}     & $45{,}211$ & $7$  & $9$ & $16$ & $0.117$ & $4{,}522$  & replication \\
\textsc{magic}\footnotemark[1]    & $19{,}019$ & $10$ & $0$ & $10$ & $0.352$ & $1{,}902$  & numerical-only control \\
\textsc{shoppers} & $12{,}330$ & $10$ & $7$ & $17$ & $0.155$ & $1{,}233$  & qualitative only \\
\botrule
\end{tabular}
\end{table}
\footnotetext[1]{The UCI listing reports $19{,}020$ rows for \textsc{magic}; the copy used throughout our
pipeline contains $19{,}019$. The difference is a single row and does not affect any reported quantity.}


\section{Reproducibility details}
\label{app:repro}

\bmhead{Discriminators}
The \xgbcst\ discriminator is a gradient-boosted-tree classifier \citep{chen2016xgboost} with $300$ estimators
at maximum depth $6$, trained with $5$-fold stratified cross-validation on a balanced (equal-size) two-sample
set; we report the out-of-fold AUC canonicalized as $\mathrm{c2st}=\max(\mathrm{AUC},1-\mathrm{AUC})$. The
discriminator-strength robustness check (Appendix~\ref{app:extra}) additionally uses $100/300/600$ estimators
at depth $3/6/10$. The linear-discriminator C2ST is a logistic-regression detection score
(SDMetrics-style \citep{patki2016sdv}), reported so that $1.0$ denotes a discriminator at chance
(AUC${\approx}0.5$); Trend is the SDMetrics pairwise column-correlation similarity.

\bmhead{Evaluation protocol}
Each dataset uses a fixed held-out test set (Table~\ref{tab:datasets}); references and generators are
evaluated against this set only. We report $\mathrm{mean}\pm\mathrm{std}$ over five seeds, which vary network initialization and data
ordering, and, with the \textsc{TabDiff} exception noted below, sampling noise as well. The
fully-factorized reference resamples each column independently from its empirical marginal; the oracle is a
fresh real-data subsample; the numerical--categorical cross component uses a block permutation that pairs the
numerical block of row $i$ with the categorical block of row $\pi(i)$. All diagnostic computations are
deterministic given the generated samples and the seed. One generator is an exception on the generation side:
\textsc{TabDiff}'s released sampler does not reseed before drawing, so repeated sampling from a single
checkpoint yields different draws, which is why its default-step baseline in the step-count sweep differs
slightly from Table~\ref{tab:tabdiff}. With five seeds, an exact sign-flip test has a minimum two-sided $p$ of $0.0625$, so significance alone is
blunt at this sample size. We report confidence intervals ($t$-distribution, four degrees of freedom) as the
primary evidence, with parametric $t$ $p$-values and sign counts as effect-size summaries; the interval and
the test share assumptions and, two-sided at the same level, the same decision, but the interval also bounds
the effect size, which is what the negative results need.

\bmhead{Generators and capacity sweep}
\tabbyflow/EF-VFM \citep{tabbyflow} and \textsc{TabDiff} \citep{tabdiff} are trained with the authors'
released code and default recipes; the second-generator check verifies that \textsc{TabDiff}'s reported
Shape-similarity and machine-learning-efficacy scores are reproduced before measuring its dependency gap
(Section~\ref{sec:moves}). The capacity sweep varies the network width $\mathrm{dim}_t\in\{1024,2048,4096\}$
($1{\times}/2{\times}/4{\times}$; $10.6$M$/42$M$/168$M parameters), holding everything else fixed, and reports
each run's final training loss as a convergence gate; \textsc{default} fails this gate at all widths (its plateau
scheduler collapses the learning rate) and is excluded from the capacity comparison. The downward sweep
(Section~\ref{sec:moves}) extends the same protocol to $\mathrm{dim}_t\in\{128,256,512\}$ on \textsc{adult}
with five matched seeds, reusing the published $1{\times}$ checkpoints as the reference point rather than
retraining them.

\bmhead{Samplers and step counts}
EF-VFM is sampled with its released default, an adaptive Dormand--Prince (\texttt{dopri5}) solver, whose
realized number of function evaluations varies by seed ($674\pm66$ on \textsc{adult}, $446\pm14$ on
\textsc{default}); \textsc{TabDiff} uses its released fixed step count together with the stochastic sampler of
\citet{tabdiff}. The step-count sweep (Section~\ref{sec:moves}) varies only the number of function evaluations, resampling
from the same trained checkpoints, over a range bracketing each generator's default; the contrast reported in
Section~\ref{sec:moves} is between the default setting and the largest step count. The complete
codebase and configuration files are covered by the Data and Code Availability statement.


\bmhead{Optimization and sampling notes referenced in Section~\ref{sec:moves}}
On \textsc{default}, optimization is indistinguishable across widths: at $1{\times}$, $2{\times}$, and $4{\times}$ alike the plateau scheduler collapses the learning rate mid-training, and all runs' final losses cluster at ${\approx}19$ ($18.8$--$19.3$ across all width-sweep runs). We therefore cannot verify that the larger models are trained to their capacity (the matched-optimization premise of the sweep), so \textsc{default} is excluded from the capacity comparison. This does not implicate the Section~\ref{sec:diagnosis} diagnosis, which evaluates the generator produced by the standard recipe against same-dataset \ff/oracle references and requires no optimization matched across widths; under-optimization is itself consistent with the H3 account.
We train \textsc{TabDiff} with the authors' released code and default recipe. On \textsc{adult} and \textsc{default}, its column-wise Shape-similarity and train-on-synthetic (machine-learning-efficacy) AUC scores match those reported by \citet{tabdiff} to within $0.05$ percentage points and $0.007$ respectively, indicating the model reaches its usual operating point, so the measured dependency gap is not an artifact of under-training. Pairwise Trend agrees on \textsc{adult}; on \textsc{default} our Trend error is \emph{lower} than reported (plausibly a data-split or SDMetrics-version difference) and does not enter the \xgbcst-based dependency measure.
The default-step baseline in this sweep ($0.051$) differs slightly from the $0.054$ of Table~\ref{tab:tabdiff} because \textsc{TabDiff}'s sampler is not reseeded before generation, so the two are different draws from the same trained checkpoints; each value lies inside the other's seed-level interval. The fraction quoted is computed within the sweep, against its own default-step baseline.


\section{Sufficiency of per-dimension means for the diffusion generator}
\label{app:diffusion}

The argument for the diffusion generator is the same in structure and rests entirely on standard results. For its continuous columns, the additive-over-dimensions denoising objective is minimized by the per-dimension posterior means $\mathbb{E}[x_0^i\mid x_t]$, which determine the score by Tweedie's formula \citep{efron2011tweedie,vincent2011connection}, and the score determines the reverse-time dynamics that transport to the true \emph{joint} \citep{song2021score}. For its masked (absorbing) categorical columns, the factorized predictor of the clean value at each masked position is exact in the continuous-time limit, where unmasking events are almost surely not simultaneous, so the joint is likewise recovered \citep{campbell2022continuous,ou2024absorbing}. At a \emph{finite} step count, however, several positions are decoded in parallel from a factorized predictive and joint structure can be lost. This discretization contribution is one the sufficiency argument does not cover for either generator; Section~\ref{sec:moves} probes it directly.


\section{Negative results in full}
\label{app:negative}

\bmhead{The explicit cross-coupling head}
The intervention produces no significant change on any of the three datasets we test. 
On \textsc{adult} the targeted component changes by $\Delta \mathrm{dep}_{\mathrm{cross}}={+}0.002\pm0.003$ (the wrong direction); on \textsc{bank} by $-0.006\pm0.006$; on \textsc{default} by $+0.003\pm0.009$.
In all cases the standard deviation is at least the magnitude of the mean, i.e., no significant change. 
Minority-class $F_1$ likewise shows no significant improvement anywhere ($\Delta_{F_1}={+}0.008/{+}0.002/{+}0.005$, all within noise), and the full \xgbcst\ shows no significant change (on \textsc{bank} the point estimates move in the favorable direction, with all five seeds' full \xgbcst\ decreasing slightly ($p=0.092$), but remain within noise).
None of these metrics meet the prespecified criteria.
To quantify what this negative result rules out, we report 95\% confidence intervals on the paired differences ($t$-distribution, 4 degrees of freedom). 
On \textsc{adult} the interval for the targeted cross component is $[-0.001,\,+0.005]$: a true improvement larger than $0.001$ (about $3\%$ of \textsc{adult}'s $0.034$ cross component) is excluded. 
On \textsc{bank} and \textsc{default} the intervals are wider ($[-0.014,\,+0.002]$ and $[-0.008,\,+0.014]$), so improvements of roughly $0.01$ cannot be excluded there; the tight null is \textsc{adult}'s.
The sampler-level check passed on all three datasets, so this is a failure of the \emph{remedy}, not of the mechanism's wiring; it shows the added pathway is active at sampling time, which is what it was built to catch, and is not a test of expressive power. 
The scope of this null, and why it is over-determined, is discussed in Section~\ref{sec:negative}.
We stress that the paired protocol only removes shared noise (training trajectories still diverge), and sign counts are screening, not significance; we therefore read the magnitudes, which are within noise. 
A single-seed preview on \textsc{adult} had looked encouraging and reversed sign once all five seeds were in, a cautionary reminder against single-run claims.

\bmhead{Post-hoc copula and resampling routes}
The standard route for imposing dependency on a set of marginals, a fitted copula, cannot supply the missing structure on these datasets, because it does not fit them in the first place.
Fitted directly to the real training data, a Gaussian copula produces samples that our discriminator separates at the $\approx\!1.0$ ceiling of \xgbcst, so it cannot serve as a post-hoc dependency injector on these datasets. We do not decompose this failure and do not attribute it to the copula's dependence model specifically: at the ceiling the marginal and dependency components are not separately identifiable, and a mixed-type schema also stresses the rank transform that the fit relies on. What the result does establish is narrow and sufficient for our purpose, namely that this route is not available here. We test only this variant, and accordingly do not claim a general negative result for richer copula families (e.g., vine copulas \citep{aas2009vine}); establishing one would require a controlled comparison we do not carry out here. 
A resampling baseline (the Synthetic Minority Over-sampling Technique, SMOTE \citep{chawla2002smote}) reaches minority-class $F_1$ comparable to training on real data (with higher recall), but by \emph{reusing} real majority rows and linearly interpolating minority samples rather than synthesizing the joint.
Its fidelity is dominated by the real component, so it neither exhibits the generator's gap nor speaks to whether a \emph{generator} can recover dependency; it shows only that good minority utility is reachable by real-data oversampling, a fact orthogonal to the generative-dependency question.


\section{Additional results}
\label{app:extra}

\bmhead{Independent categorical decoding does not lose joint structure (controlled XOR)}
The final sampling step of EF-VFM emits each categorical column from its own head. To test whether this
discards joint categorical structure, we build a controlled task in which a third categorical variable is
the exclusive-or of two others ($C_3=C_1\oplus C_2$), so that all pairwise correlations are zero while the
three-way joint is fully determined. This is precisely the kind of higher-order dependence a second-moment
view would miss. An EF-VFM-style model recovers the joint (the fraction of generated rows satisfying
$C_3=C_1\oplus C_2$, which is $1.0$ in real data) at $0.98$ with five sampling steps and $1.00$ with twenty,
\emph{despite} independent per-column decoding. Independent decoding therefore does not lose joint
categorical structure once the velocity field is accurate, consistent with the recovery argument of
Section~\ref{sec:analysis}; the residual dependence on step count is the discretization component that the
sweep on real data in Section~\ref{sec:moves} quantifies.

\bmhead{Per-dataset detail of the upward capacity sweep}
Per-seed linear trends across the three widths agree with the paired tests reported in
Section~\ref{sec:moves}: the slope (dependency per width-doubling) is $+0.004$ on \textsc{adult}
($t=2.15$), $+0.002$ on \textsc{bank} ($t=0.77$), and ${-}0.005$ on \textsc{magic} ($t={-}1.60$), none
significant at five seeds. \textsc{magic} is the weakest cell and, if anything, points against H3: its three
widths decrease monotonically ($0.025\!\to\!0.016\!\to\!0.015$, a $40\%$ drop) and its paired CI
$[-0.027,\,+0.007]$ has a lower-bound magnitude ($0.027$) exceeding its \emph{entire} dependency component
($0.025$), so the \textsc{magic} data are fully compatible with scaling closing the gap. We report this
rather than absorb it into an aggregate: the capacity conclusion rests on \textsc{adult} and, more weakly,
\textsc{bank}, and \textsc{magic}'s interval carries no diagnostic power.

\bmhead{Seed-level behavior of minority $F_1$ in the downward sweep}
Across the twenty runs of the downward sweep (four widths $\times$ five seeds on \textsc{adult}), nineteen
minority-$F_1$ values fall in $0.655$--$0.675$ and one falls at $0.617$. That run's final training loss,
marginal component, and dependency component are all in line with its width group, so we retain it rather
than treat it as a failed run. Its existence does mean that the seed-level distribution of minority $F_1$
has heavier tails than the within-width standard deviations ($\approx\!0.005$) suggest, and that the
$t$-intervals we report for this quantity assume more regularity than the data demonstrate. The
$\tfrac18{\times}$ versus $1{\times}$ contrast used in Section~\ref{sec:moves} contains no such point; the
distribution-free statement that survives regardless is that dependency varies by a factor of two across
these widths while nineteen of twenty $F_1$ values lie within a band of $0.02$.

\bmhead{Discriminator-strength robustness for the second generator}
Across three discriminator strengths (gradient-boosted trees with $100/300/600$ estimators at depth
$3/6/10$), \textsc{TabDiff}'s dependency component tracks \tabbyflow's at all settings
($\mathrm{dep}={+}0.018/{+}0.054/{+}0.066$ versus ${+}0.014/{+}0.050/{+}0.067$ on \textsc{adult}; the middle
setting is the one used throughout, matching Table~\ref{tab:diagnosis}), and the
\ff{}$\,>\,$generator$\,>\,$oracle ordering is preserved at every setting. The absolute magnitude grows with
discriminator strength, which is why every component is read against \ff\ and oracle references computed
under the \emph{same} discriminator rather than as an absolute quantity.


\end{appendices}

\end{document}